\documentclass[conference]{IEEEtran}
\IEEEoverridecommandlockouts
\usepackage{cite}
\usepackage{amsmath,amssymb,amsfonts}
\usepackage{algorithmic}
\usepackage{graphicx}
\usepackage{textcomp}
\usepackage{xcolor}
\usepackage{pifont}
\usepackage{subfig}
\usepackage{multirow}
\usepackage{arydshln}

\definecolor{darkgreen}{rgb}{0.2, 0.7, 0.2}
\definecolor{darkred}{rgb}{0.9, 0.2, 0.2}
\definecolor{darkblue}{rgb}{0.0, 0.0, 0.8}

\usepackage[a4paper, total={184mm,239mm}]{geometry}
\def\BibTeX{{\rm B\kern-.05em{\sc i\kern-.025em b}\kern-.08em
    T\kern-.1667em\lower.7ex\hbox{E}\kern-.125emX}}
\begin{document}
\title{SCALES: Boost Binary Neural Network for Image Super-Resolution with Efficient Scalings}

\author{
    Renjie Wei$^{12}$, Zechun Liu$^{5}$, Yuchen Fan$^{5}$,
    Runsheng Wang$^{234}$, Ru Huang$^{234}$, and Meng Li$^{123\dag}$
\\
\textit{$^1$Institute for Artificial Intelligence \& $^2$School of Integrated Circuits, Peking University, Beijing, China} \\
\textit{$^3$Beijing Advanced Innovation Center for Integrated Circuits, Beijing, China} \\
\textit{$^4$Institute of Electronic Design Automation, Peking University, Wuxi, China}  
\textit{$^5$Meta Inc Menlo Park CA, USA}

\thanks{
This work was supported in part by National Natural Science Foundation of China under Grant 62495102 and Grant 92464104, in part by Beijing Municipal Science and Technology Program under Grant Z241100004224015, and in part by 111 Project under Grant B18001.

$^\dag$Corresponding author.}
}

\maketitle

\begin{abstract}
Deep neural networks for image super-resolution (SR) have demonstrated superior performance.
However, the large memory and computation consumption hinders their deployment on resource-constrained devices.
Binary neural networks (BNNs), which quantize the floating point weights and activations to 1-bit
can significantly reduce the cost. 
Although BNNs for image classification have made great progress these days, 
existing BNNs for SR still suffer from a large performance gap between the FP SR networks.
To this end, we observe the activation distribution in SR networks and find much larger pixel-to-pixel, channel-to-channel, layer-to-layer, and image-to-image variation in the activation distribution than image classification networks.
However, existing BNNs for SR fail to capture these variations that contain rich information for image reconstruction, leading to inferior performance.
To address this problem, we propose SCALES, a binarization method for SR networks
that consists of the layer-wise scaling factor,
the spatial re-scaling method, and the channel-wise re-scaling method,
capturing the layer-wise, pixel-wise, and channel-wise variations efficiently in an input-dependent manner.
We evaluate our method across different network architectures and datasets.
For CNN-based SR networks, our binarization method SCALES outperforms the prior art method by 0.2dB
with fewer parameters and operations.
With SCALES, we achieve the first accurate binary Transformer-based SR network,
improving PSNR by more than 1dB compared to the baseline method.

\end{abstract}

\section{Introduction}
\label{sec:introduction}

Image super-resolution (SR) is a fundamental task in computer vision.
It aims to reconstruct high-resolution (HR) images, which have more details and high-frequency information, from low-resolution (LR) images.
In recent years, deep neural networks (DNNs) have achieved great quality in image SR including convolution neural network (CNN)-based~\cite{ledig2017photo, lim2017enhanced, zhang2018residual, tian2022image}
and Transformer-based~\cite{chen2021pre, liang2021swinir, chen2023activating} methods.
However extensive parameters and computation demands of these
SR networks hinder their deployment on resource-constrained devices.


Network quantization is always a turn-to solution to reduce memory and computation costs. 
Among these, binary neural networks (BNN) quantizing the full-precision (FP) weights and activations to 1-bit
can achieve $32\times$ memory savings and $58\times$ speed up on CPUs~\cite{rastegari2016xnor}, 
which is quite effective.
However, there is still a lack of research on BNN for SR,
compared with BNN for image classification.
On the one hand, for CNN-based SR methods, 
BNN~\cite{xin2020binarized, jiang2021training, lang2022e2fif} still suffers from large performance degradation compared with their FP counterpart.
On the other hand, for Transformer-based SR methods, 
there is no research on BNN to the best of our knowledge.





\begin{figure}[!tb]
\centering
\includegraphics[width=0.49\textwidth]{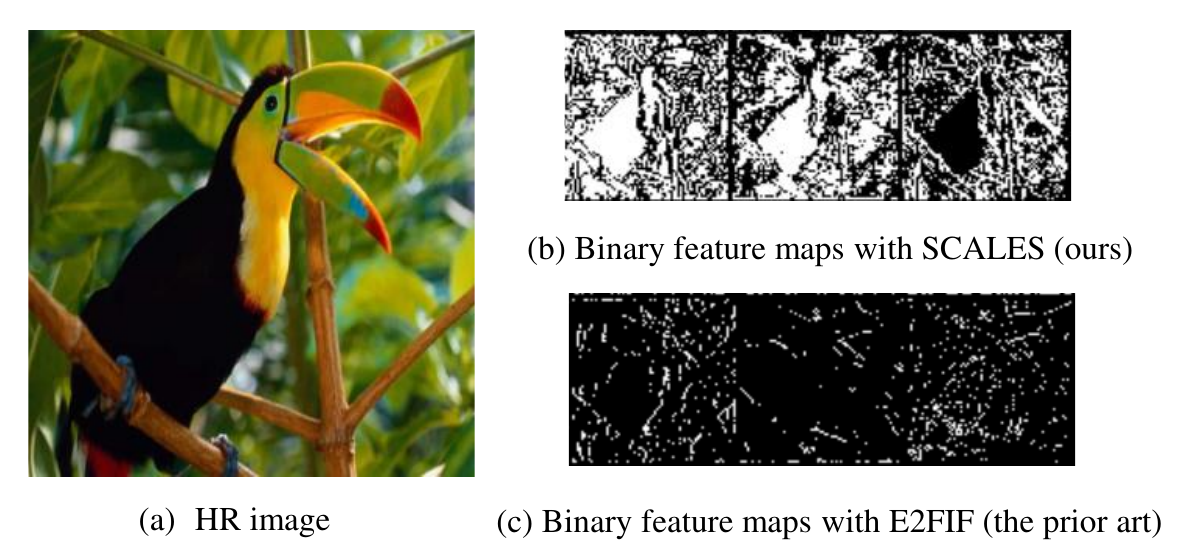}
\caption{The binary feature maps with our method SCALES and the prior art method E2FIF.} 
\vspace{-10pt}
\label{fig:intro}
\end{figure}

To this end, we study the BNN for SR comprehensively for 
both CNN-based and Transformer-based SR networks.
We discover that the activation distributions in FP SR networks including CNN and Transformer
exhibit much larger pixel-to-pixel, channel-to-channel, layer-to-layer, and image-to-image variations compared to the image classification networks.
Exiting BNNs for SR
can not capture these variations that contain detailed information for image SR
simply using the binarization methods for classification networks.
To avoid such important variations getting lost during binarization, 
we propose our method SCALES,
which consists of the layer-wise scaling factor, 
the spatial re-scaling, 
and channel-wise re-scaling method, 
to capture the layer-wise, pixel-wise, and channel-wise variations respectively in
an input-dependent manner.
With our method,
we can preserve more textures and details
for image SR compared to the prior art method in Fig.~\ref{fig:intro}.
Extensive experiments demonstrate the effectiveness of our method.
For example, for CNN-based networks,
SCALES outperforms the prior art method by 0.22dB and 0.19dB on Urban100 dataset
with less number of parameters and operations.
For Transformer-based networks,
SCALES improves PSNR by more than 1dB compared to the baseline,
leading to the first accurate binary Transformer-based SR network.

Overall, our contributions can be summarized as follows:
\begin{itemize}
    \item  We observe the activation distribution in the CNN-based and Transformer-based SR networks and discover large
    pixel-to-pixel, channel-to-channel, layer-to-layer, and image-to-image variations,
    which are important for high-performance image SR.
    \item To capture the variations, we propose a binarization method for SR networks, dubbed SCALES, which is composed of the layer-wise scaling factor, the spatial re-scaling method, 
    and the channel-wise re-scaling method.
    \item We evaluate SCALES across different SR network architectures on different benchmark datasets. 
    For CNN-based SR networks, SCALES outperforms the prior art method by 0.2dB with fewer parameters and operations. 
    With SCALES, we also achieve the first accurate binary Transformer-based SR network, improving PSNR by more than 1dB compared to the baseline method.
\end{itemize}
\section{Background}
\label{background}

\subsection{DNNs for Image SR}

DNNs have been widely used in image SR for their satisfying performance. 
SRCNN  \cite{dong2015image} first uses three convolution layers to reconstruct the HR image in an end-to-end way.
VDSR \cite{kim2016accurate} increases the network depth to 20 convolution layers and introduces global residual learning.
SRResNet \cite{ledig2017photo} introduces residual blocks as the basic block.
EDSR \cite{lim2017enhanced} removes batch normalization layers (BN) in the basic block and uses a deeper and wider model, 
which has become the standard architecture for CNN-based SR networks.  
After that, dense connect \cite{zhang2018residual}, channel attention module \cite{zhang2018image}, and non-local attention mechanism \cite{zhang2019residual} are introduced in SR networks for better image quality.
In recent years, Transformer-based SR networks
are proposed with better performance.
IPT~\cite{chen2021pre} leverages the Transformer encoder-decoder
structure and pre-training on large-scale dataset.
SwinIR~\cite{liang2021swinir} is based on Swin Transformer
and performs better than IPT.
HAT~\cite{chen2023activating} combines channel attention,
window-based self-attention, and cross-attention schemes
and reaches better image SR performance.

\begin{figure}[!tb]
\centering
\includegraphics[width=0.45\textwidth]{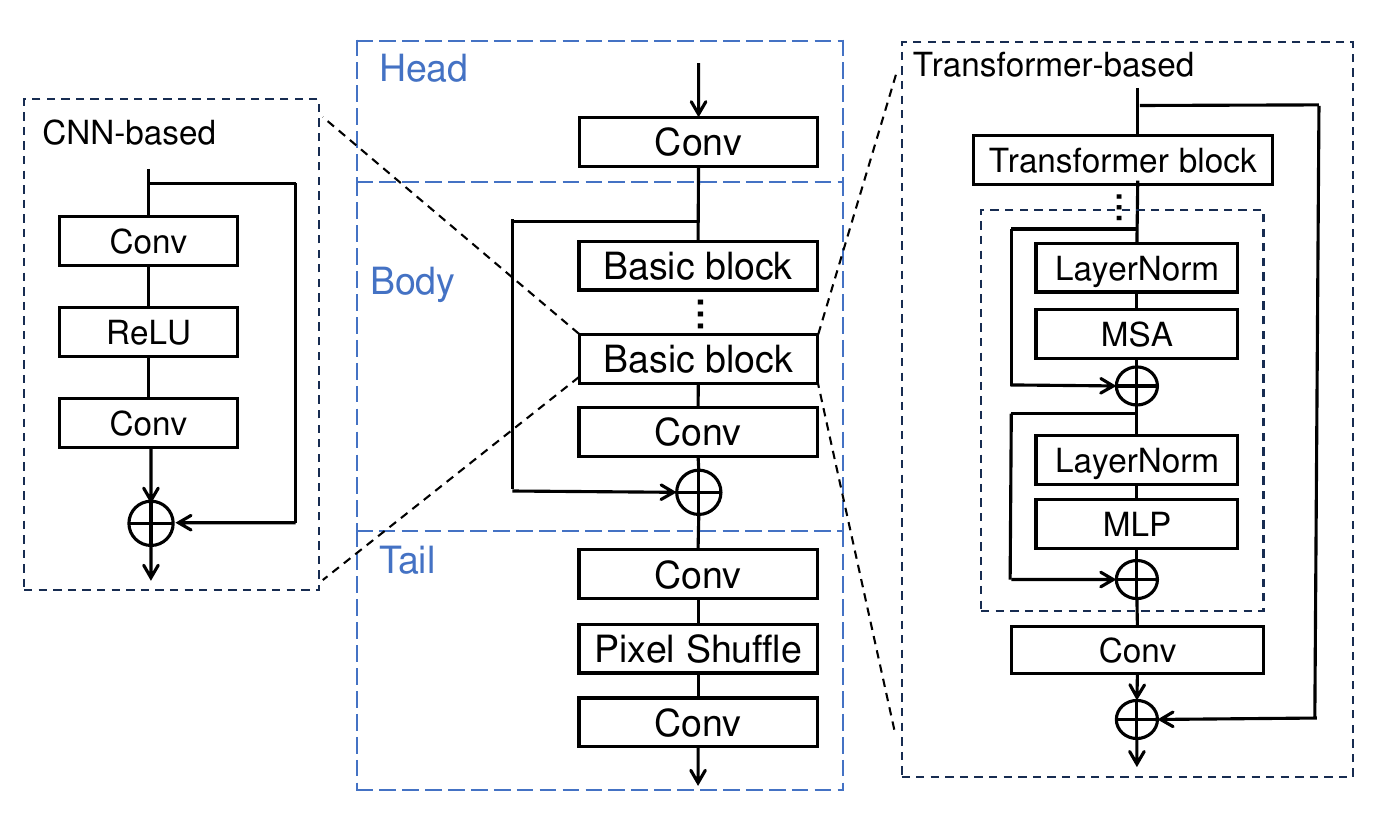}
\caption{The typical architecture of CNN-based and
Transformer-based SR networks and the detailed structure of basic blocks.
} 
\label{fig:network_arch}
\end{figure}

The architecture of typical DNNs for SR is shown in Fig.~\ref{fig:network_arch}.
It consists of three modules.
The head module extract shallow features from the input LR image.
The body module utilizes multiple basic blocks to perform deep feature extraction and the deep feature is fused with shallow feature through global residual connection.
CNN-based and Transformer-based networks
use different basic blocks as shown in Fig.~\ref{fig:network_arch}.
The former incorporates convolution layers and ReLU.
The latter incorporates transformer layers including 
layernorm, multi-head self-atteion (MSA), multi-layer perceptron (MLP) and the additional convolution layer.
The tail module reconstruct the high-resolution output by convolution and pixel shuffling.

\subsection{BNNs for SR}

BNN, which quantizes both activation and weight to $\{-1,1\}$ is first proposed by XNOR-Net \cite{rastegari2016xnor} for the image classification task.
Afterward, a lot of works\cite{lin2017towards,liu2018bi,martinez2020training,liu2020reactnet,tu2022adabin, bai2020binarybert, he2023bivit}
are proposed to reduce the classification accuracy gap between BNNs and their FP counterparts.
However, research on BNNs for SR is relatively scarce.
Table~\ref{tab:related work} lists some representative works.
It is worth noting that they are all designed for CNN-based networks.
\cite{ma2019efficient} first introduces binarization to SR networks and reduces the model size of FP SRResNet.
However, they only binarize weights and leave activations at FP, which impedes the bit-wise operation and requires expensive FP accumulations.
BAM \cite{xin2020binarized} binarizes both weights and activations using a bit-accumulation mechanism to approximate the FP convolution.
They binarize weights and activations in each layer based on the accumulation of previous layers, 
which introduces extra FP accumulation during inference.
BTM \cite{jiang2021training} finds that BN in BNNs introduces a lot of FP calculations.
They design a binary training mechanism to normalize input LR images and build a BNN without BN, named IBTM.
LMB \cite{li2022local} calculates the threshold for each pixel by averaging its neighborhood pixel values, which increases computation significantly for calculating per-pixel threshold.
DAQ \cite{hong2022daq} proposes a per-channel activation quantization method to adapt the diverse channel-wise distributions.
However, it introduces large FP computations 
for calculating the mean and standard deviation of each channel of activations.
E2FIF~\cite{lang2022e2fif} proposes an end-to-end full-precision information flow to improve the performance of BNN.
However, these methods still exhibit a performance gap compared to their FP counterparts because they fail to capture the large activation variations.
Moreover, how to binarize Transformer-based SR networks remains an unresolved issue.

\begin{table}[!tb]
\centering
\caption{Comparison between existing BNNs for SR and our work
on the spatial, channel-wise, layer-wise, image-wise adaptability,
and the hardware cost. 
}
\label{tab:related work}

\scalebox{0.7}{

\begin{tabular}{c|ccccccc}
\hline
Method                          & Spa. Adpt.    & Chl. Adpt.     & Layer Adpt.     & Img. Adpt.             & HW cost \\ \hline
\cite{ma2019efficient}          & \textcolor{darkred}{\ding{55}}     & \textcolor{darkred}{\ding{55}}      & \textcolor{darkred}{\ding{55}}       & \textcolor{darkred}{\ding{55}}            & FP Accum. \\
BAM \cite{xin2020binarized}     & \textcolor{darkgreen}{\ding{52}}     & \textcolor{darkred}{\ding{55}}      & \textcolor{darkred}{\ding{55}}       & \textcolor{darkred}{\ding{55}}            & Extra FP Accum. \\
BTM \cite{jiang2021training}    & \textcolor{darkred}{\ding{55}}     & \textcolor{darkred}{\ding{55}}      & \textcolor{darkred}{\ding{55}}       & \textcolor{darkgreen}{\ding{52}}            & Low \\
LMB \cite{li2022local}          & \textcolor{darkgreen}{\ding{52}}     & \textcolor{darkred}{\ding{55}}      & \textcolor{darkred}{\ding{55}}       & \textcolor{darkgreen}{\ding{52}}            & FP Accum. \\
DAQ \cite{hong2022daq}          & \textcolor{darkred}{\ding{55}}     & \textcolor{darkgreen}{\ding{52}}      & \textcolor{darkred}{\ding{55}}       & \textcolor{darkgreen}{\ding{52}}          & FP Mul. and Accum. \\
E2FIF \cite{lang2022e2fif}      & \textcolor{darkred}{\ding{55}}    & \textcolor{darkred}{\ding{55}}    & \textcolor{darkred}{\ding{55}}      & \textcolor{darkred}{\ding{55}}           & Low \\
\textbf{SCALES (ours)}          & \textcolor{darkgreen}{\ding{52}}    & \textcolor{darkgreen}{\ding{52}}    & \textcolor{darkgreen}{\ding{52}}       & \textcolor{darkgreen}{\ding{52}}         & Low    \\ \hline
\end{tabular}

}
\end{table}

\section{Motivation}
\label{sec:motivation}

Existing BNNs focus on binarizing the weights and the input activations of the basic blocks in the body module as shown in Fig.~\ref{fig:network_arch},
which account for most of the parameters and computations of the entire
model.
However, they ignore the large variations in activation distribution and have inferior SR performance.
In this section, we showcase that the activation distributions in FP SR networks exhibit much larger \textit{pixel-to-pixel, channel-to-channel, layer-to-layer, and image-to-image variations} than those in the image classification networks.

\subsection{Variations in CNN-based SR Network}

\begin{figure}[!tb]%
  \centering
    \subfloat[Distribution across pixels (img1)]{
        \label{subfig:CNN Distribution across pixels image1}
        \includegraphics[width=0.22\textwidth]{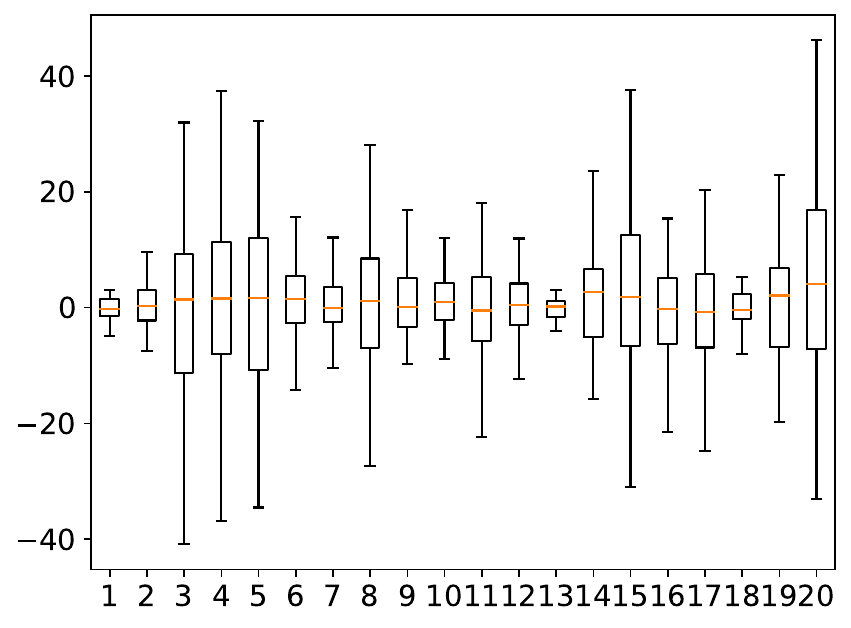}
      }
      \subfloat[Distribution across pixels (img2)]{
        \label{subfig:CNN Distribution across pixels img2}
        \includegraphics[width=0.22\textwidth]{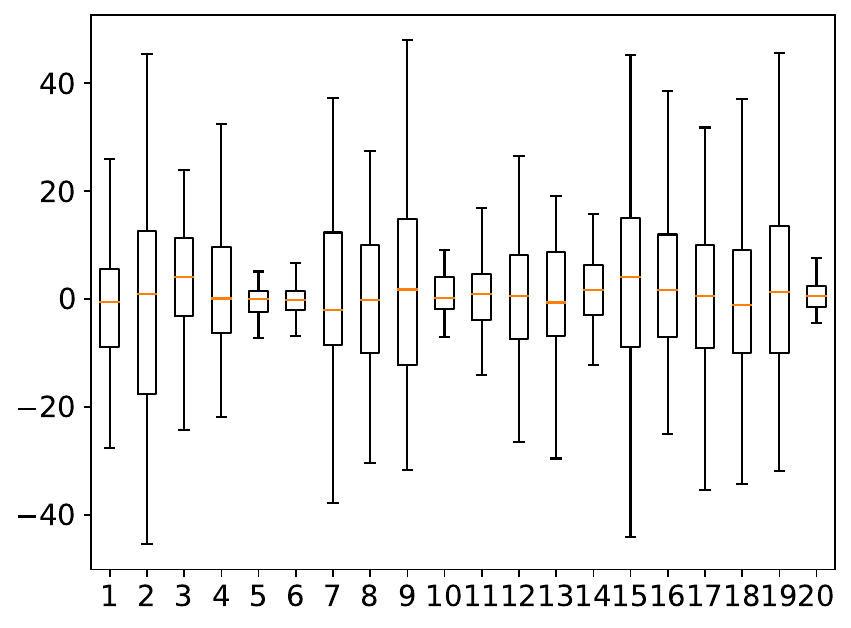}
      }
      
    \subfloat[Distribution across layers]{
        \label{subfig:CNN Distribution across layers}
        \includegraphics[width=0.22\textwidth]{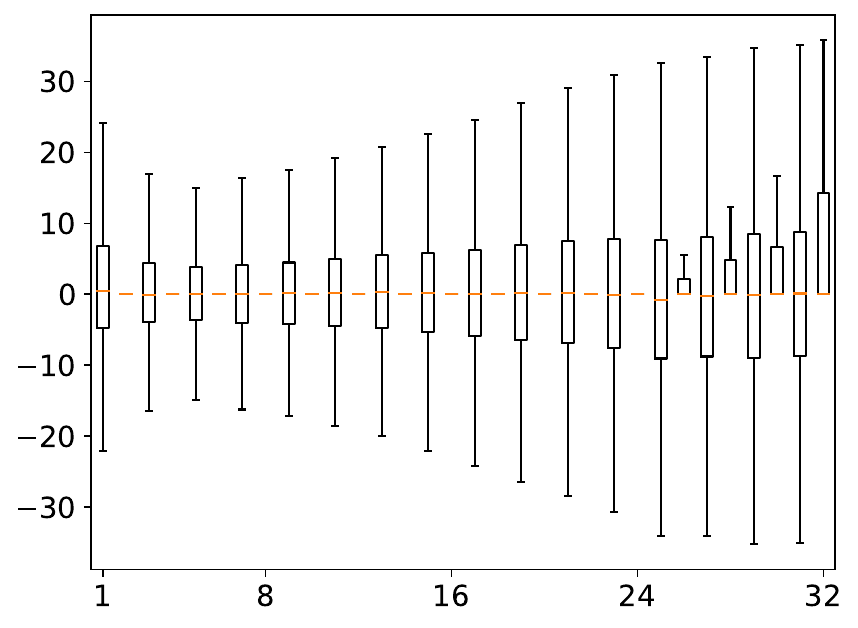}
    }
    \subfloat[Distribution across channels]{
        \label{subfig:CNN Distribution across channels}
        \includegraphics[width=0.22\textwidth]{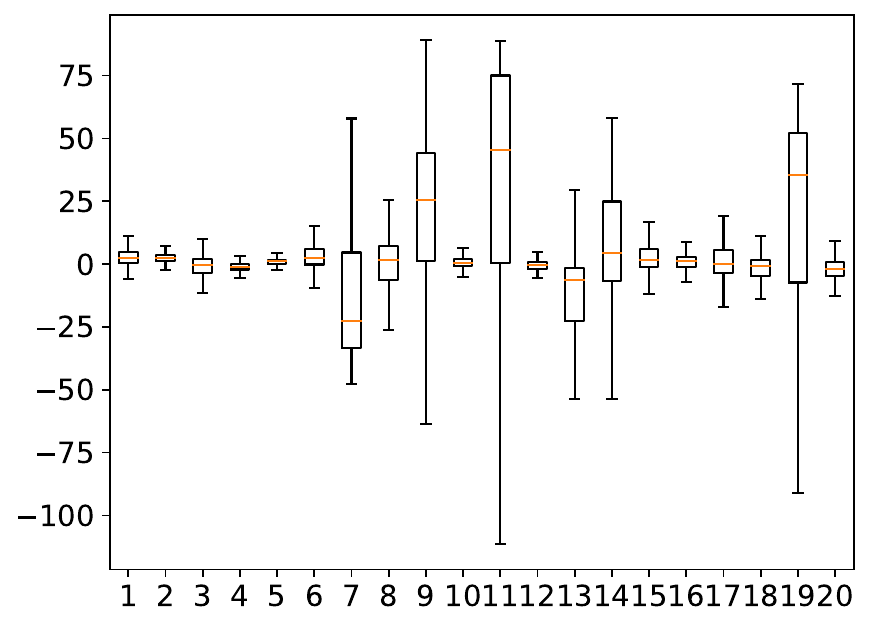}
    }
  
  \caption{
Activation distribution in EDSR~\cite{lim2017enhanced}. 
}
  \label{fig:chl}
   \vspace{-5pt} 
\end{figure}

\begin{figure}[!tb]%
  \centering
    \subfloat[Distribution in ResNet18]{
        \label{subfig:Distribution in ResNet18}
        \includegraphics[width=0.22\textwidth]{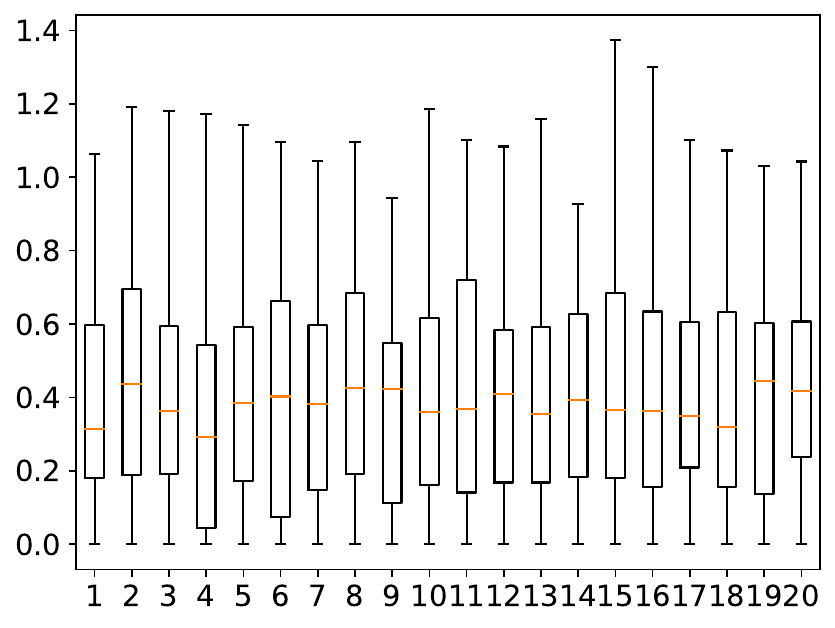}
      }
      \subfloat[Distribution in SwinViT]{
        \label{subfig:Distribution in SwinViT}
        \includegraphics[width=0.22\textwidth]{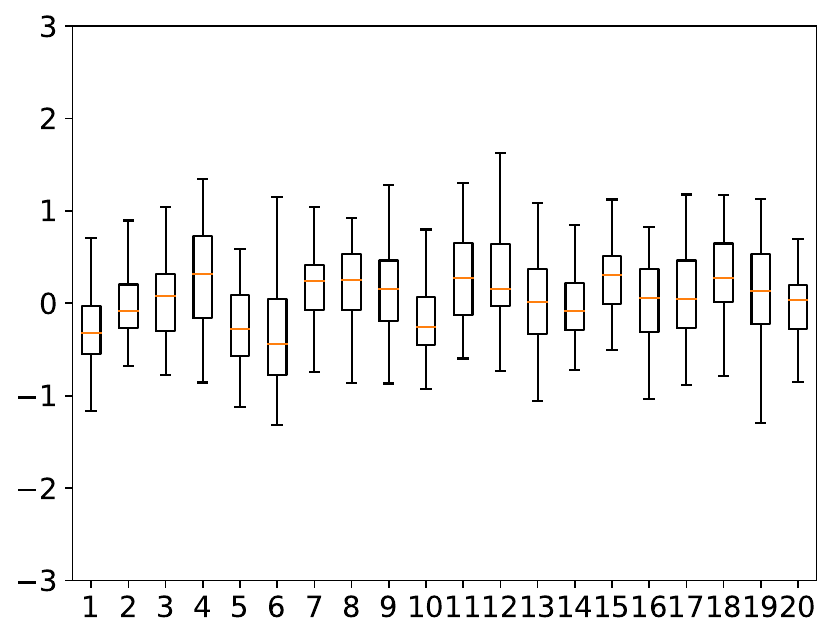}
      }
      
  \caption{
Activation distribution in CNN-based and Transformer-based classification networks ResNet18~\cite{he2016deep} and SwinViT~\cite{liu2021swin}.
The distributions across pixels, channels, layers, and images are similar, 
thus we only show distributions across pixels here.
}
  \label{fig:chl}
   \vspace{-5pt} 
\end{figure}

As shown in Fig.~\ref{subfig:CNN Distribution across pixels image1}, we random sample 20 pixels from a feature map in a CNN-based SR network EDSR,
where each pixel contains C (the number of channels) elements.
We observe large \textit{pixel-to-pixel variation},
compared to ResNet18 in Fig.~\ref{subfig:Distribution in ResNet18}.
The same holds true for \textit{channel-to-channel variation}.
The main reason is that modern SR networks removes BN
for better SR performance,
leading to large activation variations.
According to~\cite{bulat2019xnor},
the difference in activation magnitudes indicates different scaling
factors are needed. 
However, per-pixel or per-channel quantization for activation
is infeasible because they will introduce large computation overhead~\cite{xiao2023smoothquant} while existing per-tensor binarization schemes cannot capture the
variation of activation distribution.

For different layers, activations in
EDSR also exhibit large \textit{layer-to-layer variation} in Fig.~\ref{subfig:CNN Distribution across layers}.
Moreover, we find that the activations in the even layers are small, whereas the activations in the odd layers exhibit large magnitudes.
This is because for the basic block in Fig.~\ref{fig:network_arch}, the shortcut
maintains the original information of the input LR image, while the inner branch intends to re-construct the small
difference between the LR and HR image. 
Thus, the input of the first conv layer has
large magnitude, while the input of the second conv
layer has small magnitude,
which implies that different binarization schemes should be employed for different layers.
Comparing Fig.~\ref{subfig:CNN Distribution across pixels image1}
and~\ref{subfig:CNN Distribution across pixels img2},
we can also find the large \textit{image-to-image variation},
which motivates us to quantize the SR network in an input-dependent manner.

\subsection{Variations in Transformer-based SR Network}

\begin{figure}[!tb]%
  \centering
    \subfloat[Distribution across pixels (img1)]{
        \label{subfig:Distribution across pixels image1}
        \includegraphics[width=0.22\textwidth]{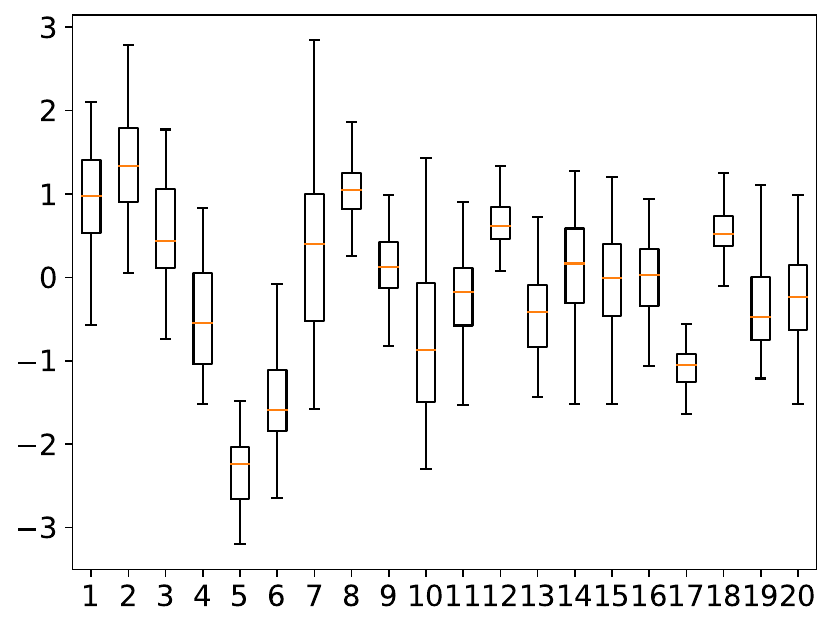}
      }
      \subfloat[Distribution across pixels (img2)]{
        \label{subfig:Distribution across pixels img2}
        \includegraphics[width=0.22\textwidth]{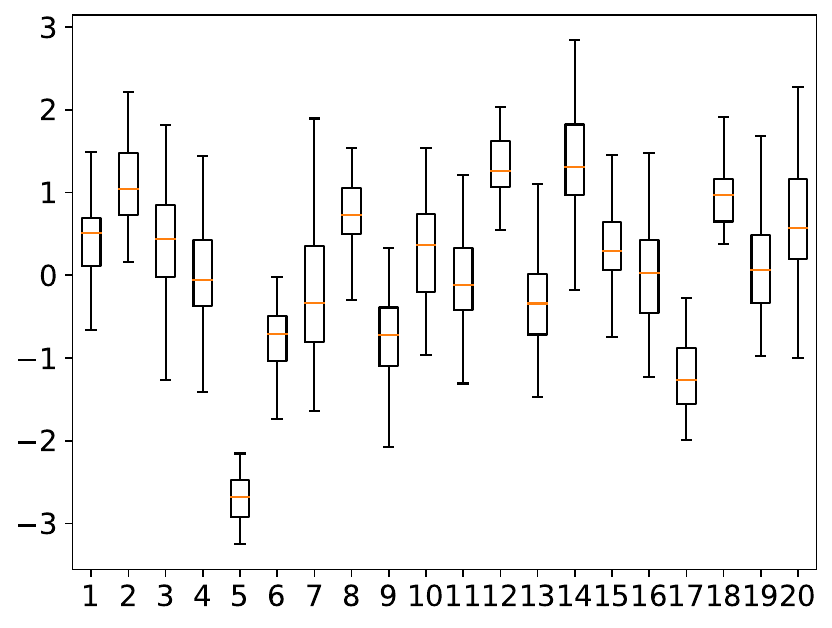}
      }
      
    \subfloat[Distribution across layers (linear)]{
        \label{subfig:Distribution across layers (linear)}
        \includegraphics[width=0.22\textwidth]{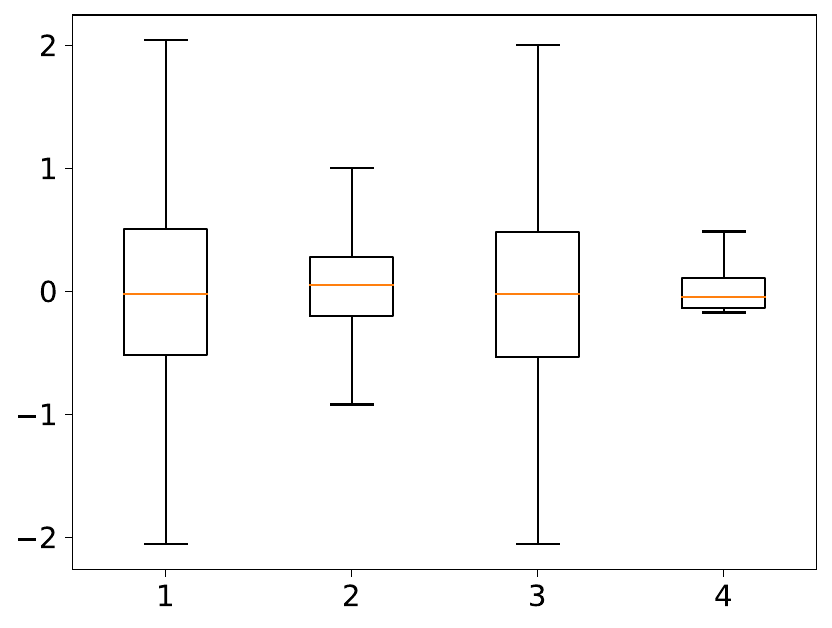}
    }
    \subfloat[Distribution across layers (conv)]{
        \label{subfig:Distribution across layers (conv)}
        \includegraphics[width=0.22\textwidth]{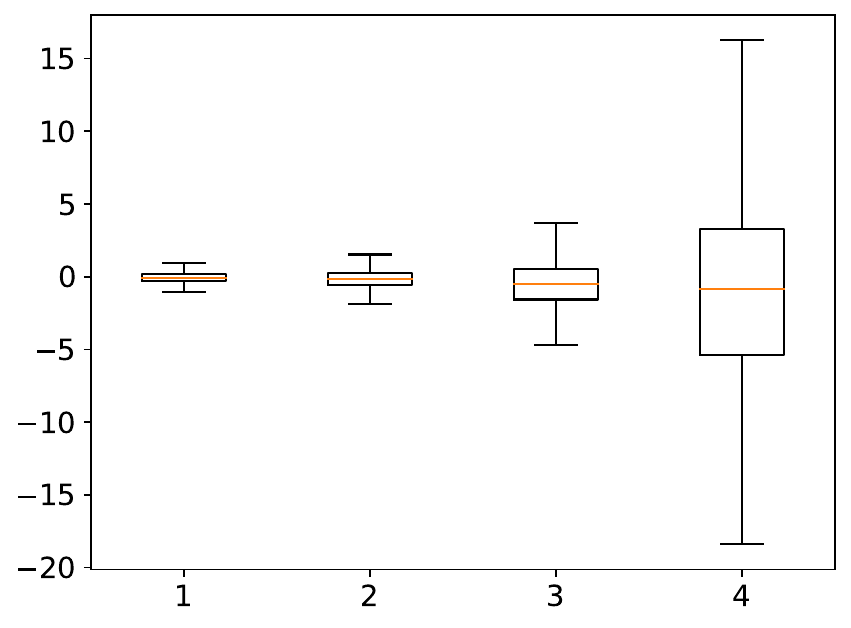}
    }
  
  \caption{
Activation distribution in SwinIR~\cite{liang2021swinir}.
}
  \label{fig:distribution in swinIR}
   \vspace{-5pt} 
\end{figure}

Large \textit{pixel-to-pixel, layer-to-layer} and \textit{image-to-image variations}
also exist in Transformer-based SR networks
as shown in Fig.~\ref{fig:distribution in swinIR}.
For the layer-to-layer variation, we plot the input activations of the four linear layers in the Transformer block
in Fig.~\ref{subfig:Distribution across layers (linear)}
and the last conv layer of each basic block in Fig.~\ref{subfig:Distribution across layers (conv)}.
It is worth noting that channel-to-channel variation 
does not exist in SwinIR,
since LayerNorm (LN) normalizes each token across the channel dimension.
Thus, for the Transformer-based SR network,
we should apply different quantization schemes
to capture the pixel-wise, layer-wise, and image-wise variations.
For quantitative comparison,
we calculate the variance of image SR and classification networks in Table~\ref{tab:variance_comparison},
which is consistent with our observations above. 

\begin{table}[!tb]
\centering
\caption{Activation variance comparison.
} 
\label{tab:variance_comparison}
\scalebox{1.0}{
\begin{tabular}{c|cc|cc}
\hline
               & EDSR    & ResNet & SwinIR & SwinViT \\ \hline
chl-to-chl     & 439.17  & 0.10   & 0.11   & 0.10    \\
pixel-to-pixel & 622.25  & 0.34   & 0.87   & 0.12    \\
layer-to-layer & 3494.38 & 0.92   & 162.70 & 3.46    \\
image-to-image & 599.39  & 0.32   & 0.84   & 0.13    \\ \hline
\end{tabular}
}
\vspace{-5pt}
\end{table}

\section{Method}
\label{sec: method}

In this section, we introduce our proposed method SCALES,
which mainly consists of three components, including the layer-wise scaling factor, the spatial re-scaling
module, and the channel-wise re-scaling module capturing the
layer-to-layer, pixel-to-pixel, and channel-to-channel variations efficiently in an input-dependent manner.

\subsection{Layer-wise Scaling Factor}
\label{subsec: layer-wise scaling factor}

Currently, existing BNNs for SR either use complicated binarization functions \cite{woo2018cbam,li2022local,hong2022daq} which have expensive computation costs, 
or use the sign function for activation binarization \cite{lang2022e2fif}, i.e., $\hat{x}=\operatorname{sign}(x)$ which can not capture the variations in SR networks.
To this end,
we propose using the layer-wise scaling factor $\alpha$
to capture the layer-to-layer variation.
Each convolution or linear layer has a scaling factor,
which is learned to have different magnitudes for different layers.
We further introduce the channel-wise threshold $\beta$ that is also learnable inspired by ReActNet \cite{liu2020reactnet}
to capture the channel-wise shifting in Fig.~\ref{subfig:CNN Distribution across channels}.
Thus, our activation binarization function becomes: 
\begin{equation}
\hat{x}=\alpha\operatorname{sign}(\frac{x-\beta}{\alpha}),
\label{eq:act_binarize}
\end{equation} 
where $\beta$ is the channel-wise learnable threshold, and $\alpha$ is the layer-wise scaling factor.
Both can be optimized end-to-end with the gradient-based method with the help of
the straight through estimator (STE)~\cite{liu2018bi}.
The gradient w.r.t. $\alpha$ can be calculated as:
\begin{small}
\begin{equation}
\begin{aligned}
&\frac{\partial \hat{x}}{\partial \alpha}=\begin{cases}-1, & \text { if } x \leq \beta-\alpha \\
-2\left(\frac{x-\beta}{\alpha}\right)^2-2 \frac{x-\beta}{\alpha}-1, &\text { if } \beta-\alpha<x \leq \beta \\
2\left(\frac{x-\beta}{\alpha}\right)^2-2 \frac{x-\beta}{\alpha}+1, &\text { if } \beta<x \leq \beta+\alpha \\
1, &\text { if } x>\beta+\alpha 
\end{cases}\\
\label{eq:grad_wrt_alpha} 
\end{aligned}
\end{equation}
\end{small}
while the gradient w.r.t. $\beta$ can be computed as: 
\begin{small}
\begin{equation}
\begin{aligned}
&\frac{\partial \hat{x}}{\partial \beta}= \begin{cases}-2-2 \frac{x-\beta}{\alpha}, & \text { if } \quad \beta-\alpha<x \leqslant \beta \\ -2+2 \frac{x-\beta}{\alpha,}, & \text { if } \beta<x \leqslant \beta+\alpha \\ 0, & \text { otherwise }\end{cases}
\label{eq:grad_wrt_beta} 
\end{aligned}
\end{equation}
\end{small}

For weights, we binarize them in a per-channel way as usual:
$\hat{w}=\frac{\|w\|_{l1}}{n} \operatorname{sign}(w),$
where $n$ denotes the number of weights.
The scaling factor is the absolute mean value for each output channel.


\subsection{Spatial Re-scaling}


\begin{figure}[!tb]%
  \centering 
  \subfloat[]{
    \label{subfig: spatial_rescale_CNN}
    \includegraphics[width=0.25\textwidth]{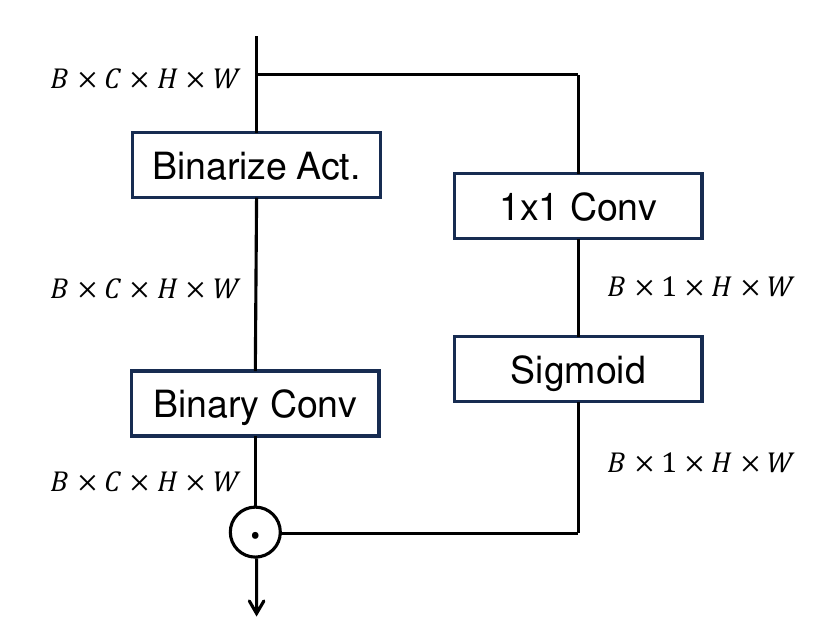}
  }
  \subfloat[]{
    \label{subfig: spatial_rescale_Transformer}
    \includegraphics[width=0.23\textwidth]{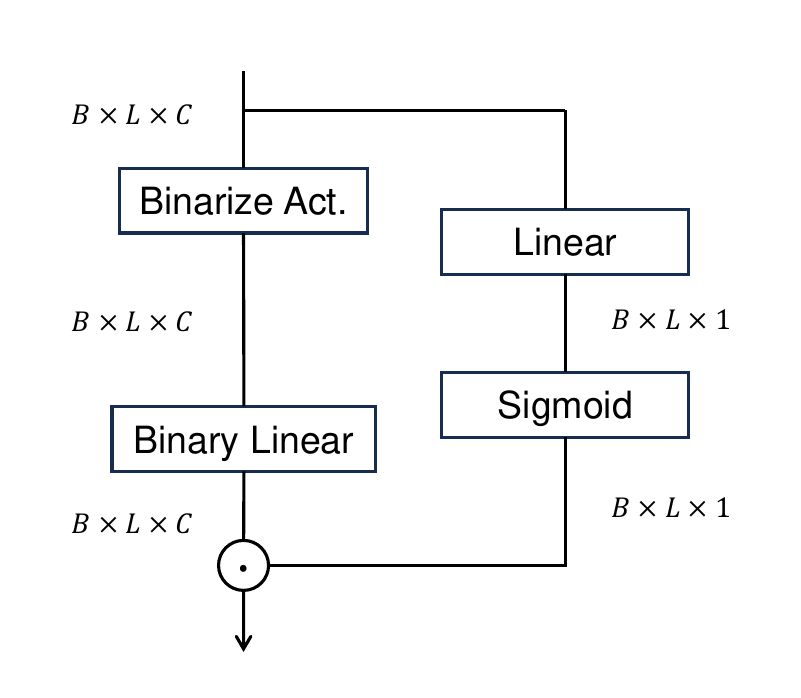}
  }
  \caption{
The proposed spatial re-scaling method for (a) CNN-based and (b) Transformer-based SR network.
}
  \label{fig:spatial_rescale}
\end{figure}

To capture the pixel-to-pixel variation in SR networks,
we propose the spatial re-scaling method for both CNN-based networks
in Fig.~\ref{subfig: spatial_rescale_CNN} and Transformer-based networks in Fig.~\ref{subfig: spatial_rescale_Transformer}.

We use the FP activation before binarization as the input
and predict the spatial scaling factors to re-scale the output of binary convolution or binary linear layer.
The spatial scaling factors are predicted
through the right-hand side branch,
i.e., the FP convolution layer with $1\times1$ kernel and sigmoid layer in Fig.~\ref{subfig: spatial_rescale_CNN}
and the FP linear layer and sigmoid layer in Fig.~\ref{subfig: spatial_rescale_Transformer}.
Although they are in FP,
they only introduce little parameters compared to the original parameters
of binary conv or linear layer.
It is also worth noting that during inference the spatial scaling factor
is not fixed but inferred from data.
Thus the spatial re-scaling module can capture spatial information
in an input-dependent manner.
As a result, the pixel-to-pixel and image-to-image variations 
are well captured.
Our spatial re-scaling method is formulated as:
\begin{equation}
A \otimes W \approx(\mathcal{B}_1(A) \otimes \mathcal{B}_2(W)) \odot S(A)
\label{eq:spatial rescale}
\end{equation}
where $A$ and $W$ are the FP activations and weights, $\mathcal{B}_1(\cdot)$ and $\mathcal{B}_2(\cdot)$ denote the binarization function for activations and weights respectively, 
$S\left(A\right)$ is our spatial re-scaling method, 
$\otimes$ denotes 
the binary convolution or multiplication operation, and 
$\odot$ denotes the broadcast element-wise multiplication.

\subsection{Channel-wise Re-scaling}

\begin{figure}[!tb]
  \centering
    \includegraphics[width=0.25\textwidth]{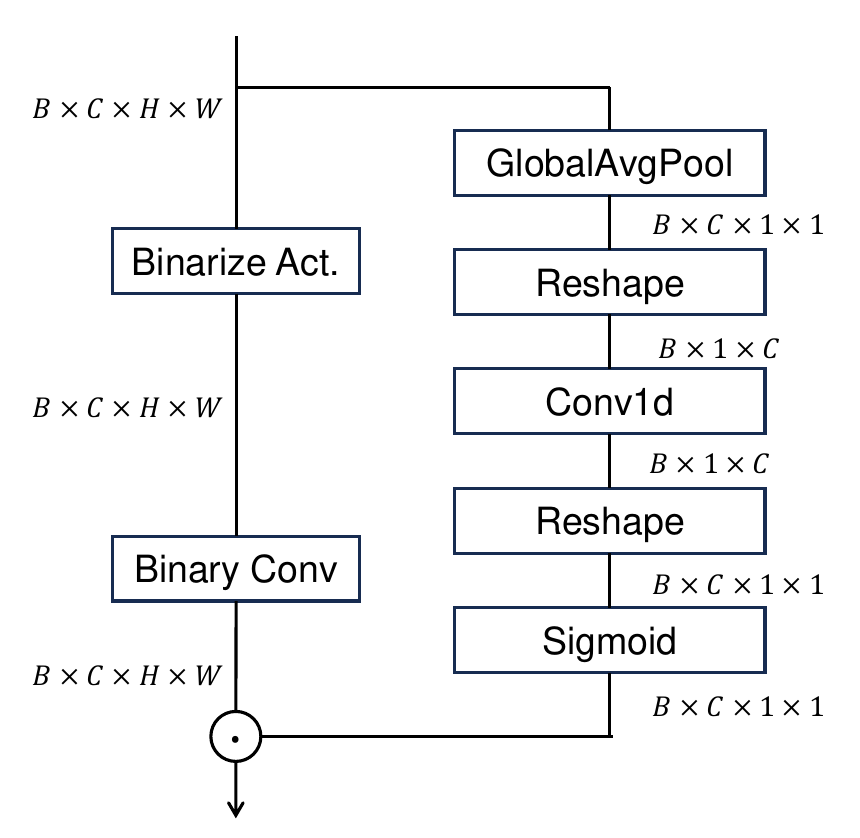}
  \caption{The proposed channel-wise re-scaling method for CNN-based SR network.} 
  \label{fig:chl_rescale}
\end{figure}

To capture the channel-to-channel variation in CNN-based SR networks,
we propose the channel-wise re-scaling method in Fig,~\ref{fig:chl_rescale}.
Note that Transformer-based SR networks do not have channel-to-channel variation due to LN.
We use the FP activations before binarization as input
since it contains rich information.
Then, a global average pooling layer is applied to aggregate the spatial information.
Afterward, we capture the inter-channel information with the Conv1d layer
and derive the channel-wise scaling factor through a sigmoid function.
Our channel-wise re-scaling method can be formulated as:
\begin{equation}
A \otimes W \approx(\mathcal{B}_1(A) \otimes \mathcal{B}_2(W)) \odot C(A)
\label{eq:channel_rescale}
\end{equation}
where $C(A)$ is our channel-wise re-scaling method.
The kernel size of the Conv1d layer is set to 5,
which we find have better performance empirically.

Previous work \cite{martinez2020training} also introduced a channel re-scaling module for image classification BNN.
However, our method differs from theirs in the way of generating the channel-wise scaling factor.
They adopt a GlobalAvgPool-Linear-ReLU-Linear-Sigmoid structure, which introduces large parameter overhead.
Our method only have $k$ FP parameters, which is the kernel size of the Conv1d layer.
While the Lineaer-ReLU-Linear structure in~\cite{martinez2020training} introduces ${2C^2} / {r}$ FP parameters, where $r$ is the compression ratio,
which are $2C^2 / rk$  times larger than ours.
The ratio will reach 1638 when $r$ is 16, $C$ is 256, and $k$ is 5, which are the typical values.

\begin{figure}[!tb]%
  \centering 
  \subfloat[]{
    \label{subfig:conv_module}
    \includegraphics[width=0.26\textwidth]{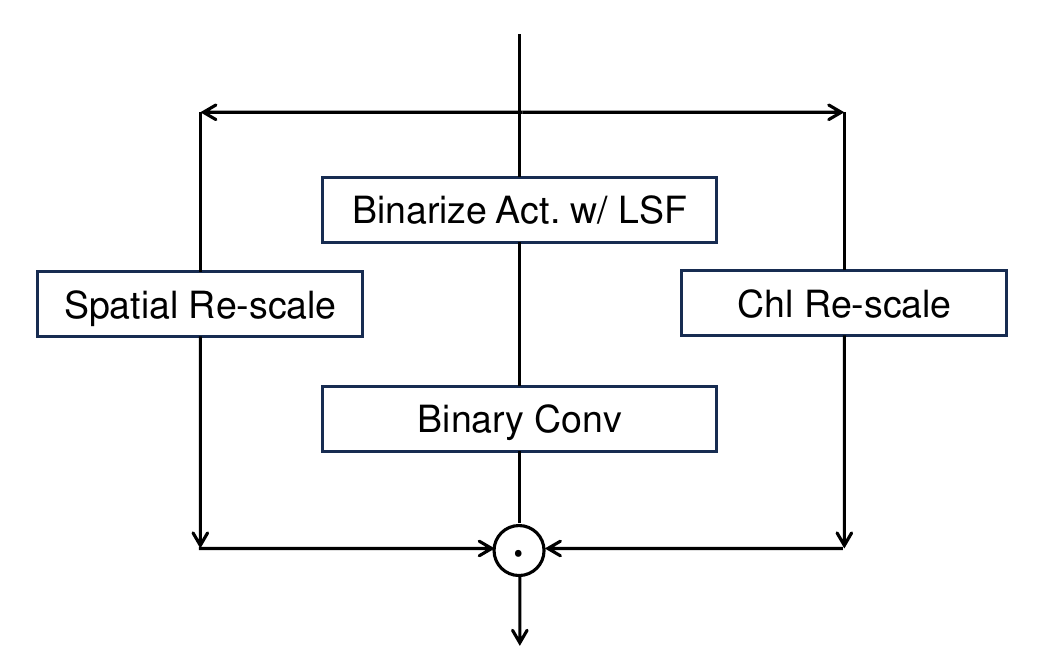}
  }
  \subfloat[]{
    \label{subfig:linear_module}
    \includegraphics[width=0.2\textwidth]{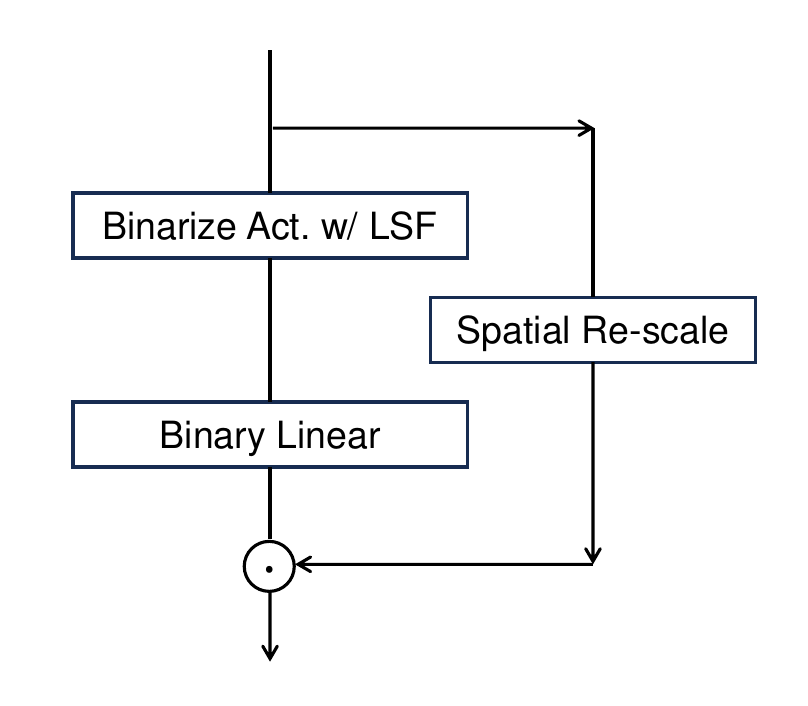}
  }
  \caption{
The binary (a) convolution and (b) linear layer integrated with SCALES. LSF stands for the layer-wise scaling factor.
}
  \label{fig:layer_with_SCALES}
\end{figure}

Combining the aforementioned methods
we achieve our proposed binarization method, SCALES. 
Fig.~\ref{fig:layer_with_SCALES} shows the binary convolution and linear layer equipped with SCALES.
For convolution, we also incorporate a skip connection following~\cite{liu2018bi,lang2022e2fif}.
The binary convolution and linear layers can serve as a drop-in replacement for various SR network architectures,
which enable the binarized SR network to efficiently capture the important
pixel-to-pixel, channel-to-channel, layer-to-layer, and image-to-image variations.


\section{Experiments}
\label{sec:experiments}

\subsection{Network Architectures}
We evaluate our proposed method SCALES on different SR network architectures.
For CNN-based SR networks,
we choose SRResNet \cite{ledig2017photo}, EDSR \cite{lim2017enhanced}, RDN \cite{zhang2018residual}, and RCAN \cite{zhang2018image}.
For Transformer-based SR networks,
we choose SwinIR (Lightweight)~\cite{liang2021swinir} and HAT~\cite{chen2023activating}.
Following existing works, the head and tail modules are not binarized.

\subsection{Experimental Settings}
We train all the models on the training set of DIV2K \cite{timofte2017ntire}. 
For evaluation, we use four standard benchmark datasets including Set5 \cite{bevilacqua2012low}, Set14 \cite{zeyde2012single}, B100 \cite{martin2001database} and Urban100 \cite{huang2015single}. 
For evaluation metrics, we use PSNR and SSIM \cite{wang2004image} over the Y channel of transformed YCbCr space.
We choose L1 loss between the SR image and the HR image as our loss function. 
Input patch size is set to $48 \times 48$. The batch size is set to 16. We use ADAM optimizer with $\beta_1=0.9$, $\beta_2=0.999$, and $\epsilon=10^{-8}$. 
We train our models for 300 epochs from scratch.
The learning rate is initialized as $2 \times 10^{-4}$ and halved every 200 epochs.

\subsection{Quantitative and Qualitative Results}

\begin{table*}[!tb]
\centering
\caption{Comparison of different methods on CNN-based SR network (SRResNet).}
\label{tab:result_srresnet}

\scalebox{1.0}{

\begin{tabular}{l|c|c|c|cc|cc|cc|cc}
\hline
\multirow{2}{*}{Method} & \multirow{2}{*}{Scale} & \multirow{2}{*}{Params} & \multirow{2}{*}{OPs} & \multicolumn{2}{c|}{Set5}       & \multicolumn{2}{c|}{Set14}      & \multicolumn{2}{c|}{B100}       & \multicolumn{2}{c}{Urban100}    \\ \cline{5-12} 
                        &                        &                         &                      & PSNR           & SSIM           & PSNR           & SSIM           & PSNR           & SSIM           & PSNR           & SSIM           \\ \hline
SRResNet-FP             & x2                     & 1517K                   & 913.8G               & 37.76          & 0.958          & 33.27          & 0.914          & 31.95          & 0.895          & 31.28          & 0.919          \\
Bicubic                 & x2                     & -                       & -                    & 33.66          & 0.930          & 30.24          & 0.869          & 29.56          & 0.843          & 26.88          & 0.840          \\
SRResNet-BAM            & x2                     & 37K                     & 28.5G                & 37.21          & 0.956          & 32.74          & 0.910          & 31.60          & 0.891          & 30.20          & 0.906          \\
SRResNet-BTM            & x2                     & 35K                     & 25.8G                & 37.22          & 0.957          & 32.93          & 0.912          & 31.77          & 0.894          & 30.79          & 0.914          \\
SRResNet-E2FIF          & x2                     & 35K                     & 25.8G                & 37.50          & \textbf{0.958} & 32.96          & 0.911          & 31.79          & 0.894          & 30.73          & 0.913          \\
SRResNet-SCALES (ours)  & x2                     & 34K                     & 24.5G                & \textbf{37.56} & \textbf{0.958} & \textbf{33.10} & \textbf{0.912} & \textbf{31.83} & \textbf{0.895} & \textbf{30.95} & \textbf{0.915} \\ \hline
SRResNet-FP             & x4                     & 1517K                   & 228.5G               & 31.76          & 0.888          & 28.25          & 0.773          & 27.38          & 0.727          & 25.54          & 0.767          \\
Bicubic                 & x4                     & -                       & -                    & 28.42          & 0.810          & 26.00          & 0.703          & 25.96          & 0.668          & 23.14          & 0.658          \\
SRResNet-BAM            & x4                     & 37K                     & 7.1G                 & 31.24          & 0.878          & 27.97          & 0.765          & 27.15          & 0.719          & 24.95          & 0.745          \\
SRResNet-BTM            & x4                     & 35K                     & 6.4G                 & 31.25          & 0.878          & 27.94          & 0.765          & 27.18          & 0.720          & 25.01          & 0.748          \\
SRResNet-E2FIF          & x4                     & 35K                     & 6.4G                 & 31.33          & 0.880          & 27.93          & 0.766          & 27.20          & 0.723          & 25.08          & 0.750          \\
SRResNet-SCALES (ours)  & x4                     & 34K                     & 6.1G                 & \textbf{31.54} & \textbf{0.883} & \textbf{28.15} & \textbf{0.770} & \textbf{27.28} & \textbf{0.726} & \textbf{25.27} & \textbf{0.757} \\ \hline
\end{tabular}

}
\end{table*}

\begin{table*}[!tb]
\centering
\caption{Comparison of different methods on Transformer-based SR network (SwinIR and HAT).}
\label{tab:result_swinir_and_hat}

\scalebox{1.0}{

\begin{tabular}{l|c|c|c|cc|cc|cc|cc}
\hline
\multirow{2}{*}{Method} & \multirow{2}{*}{Scale} & \multirow{2}{*}{Params} & \multirow{2}{*}{OPs} & \multicolumn{2}{c|}{Set5}       & \multicolumn{2}{c|}{Set14}      & \multicolumn{2}{c|}{B100}       & \multicolumn{2}{c}{Urban100}    \\ \cline{5-12} 
                        &                        &                         &                      & PSNR           & SSIM           & PSNR           & SSIM           & PSNR           & SSIM           & PSNR           & SSIM           \\ \hline
SwinIR-FP               & x2                     & 878K                    & 391.2G               & 38.14          & 0.961          & 33.86          & 0.921          & 32.31          & 0.901          & 32.76          & 0.934          \\
SwinIR-BiBERT           & x2                     & 66K                     & 12.5G                & 35.58          & 0.947          & 31.79          & 0.900          & 30.80          & 0.880          & 28.34          & 0.877          \\
SwinIR-SCALES (ours)    & x2                     & 73K                     & 15.3G                & \textbf{36.97} & \textbf{0.956} & \textbf{32.53} & \textbf{0.908} & \textbf{31.39} & \textbf{0.889} & \textbf{29.56} & \textbf{0.897} \\ \hline
SwinIR-FP               & x4                     & 897K                    & 99.2G                & 32.44          & 0.898          & 28.77          & 0.786          & 27.69          & 0.741          & 26.47          & 0.798          \\
SwinIR-BiBERT           & x4                     & 86K                     & 3.2G                 & 29.52          & 0.835          & 26.80          & 0.734          & 26.50          & 0.697          & 23.77          & 0.690          \\
SwinIR-SCALES (ours)    & x4                     & 93K                     & 3.9G                 & \textbf{29.96} & \textbf{0.849} & \textbf{27.13} & \textbf{0.743} & \textbf{26.67} & \textbf{0.704} & \textbf{24.06} & \textbf{0.704} \\ \hline\hline
HAT-FP                  & x2                     & 20.44M                  & 807.6G               & 38.73          & 0.964          & 35.13          & 0.928          & 32.69          & 0.906          & 34.81          & 0.949          \\
HAT-BiBERT              & x2                     & 0.86M                   & 25.8G                & 28.29          & 0.793          & 26.46          & 0.722          & 26.46          & 0.699          & 24.13          & 0.698          \\
HAT-SCALES (ours)       & x2                     & 0.91M                   & 35.9G                & \textbf{37.34} & \textbf{0.958} & \textbf{32.97} & \textbf{0.912} & \textbf{31.76} & \textbf{0.894} & \textbf{30.61} & \textbf{0.912} \\ \hline
HAT-FP                  & x4                     & 20.80M                  & 204.8G               & 33.18          & 0.907          & 29.38          & 0.800          & 28.05          & 0.753          & 28.37          & 0.845          \\
HAT-BiBERT              & x4                     & 1.01M                   & 6.6G                 & 26.92          & 0.774          & 25.02          & 0.671          & 25.23          & 0.645          & 22.65          & 0.639          \\
HAT-SCALES (ours)       & x4                     & 1.06M                   & 9.3G                 & \textbf{31.23} & \textbf{0.881} & \textbf{27.96} & \textbf{0.766} & \textbf{27.17} & \textbf{0.722} & \textbf{24.98} & \textbf{0.747} \\ \hline
\end{tabular}

}
\end{table*}

We evaluate SCALES on different network architectures on four benchmark datasets across different scales.
Due to page limitation, we only show the results on SRResNet, SwinIR, and HAT at $\times 2$ and $\times 4$ scale. 
For CNN-based SR networks,
as shown in Table~\ref{tab:result_srresnet},
SCALES outperforms the other methods.
It surpasses the prior art method E2FIF~\cite{lang2022e2fif},
by 0.22dB and 0.19dB on Urban100 at $\times2$ and $\times4$ scale, respectively.
For Transformer-based SR networks,
since there is no existing research on binarization,
we build the baseline model leveraging the binarization method 
proposed in BiBERT~\cite{bai2020binarybert}.
We also try the method in BiViT~\cite{he2023bivit},
but find it less effective than BiBERT~\cite{bai2020binarybert},
thus we choose the better one as our baseline.
As shown in Table~\ref{tab:result_swinir_and_hat},
our proposed method SCALES
significantly surpasses the baseline.
For example, on SwinIR,
SCALES achieves 1.39dB improvement over the baseline 
on Set5 at $\times2$ scale.
On HAT,
the improvement with our method is even larger,
i.e., 1.94$\sim$4.31dB across four datasets at $\times4$ scale.
Through our method,
we achieve the first accurate binary SR Transformer.

\begin{figure}[!tb]%
  \centering 
  \hspace*{-0.5cm}
  \subfloat[Visual results from Urban100 at $\times 4$ scale on RCAN architecture]{
    \label{subfig:visual_result_RCAN}
    \includegraphics[width=0.45\textwidth]{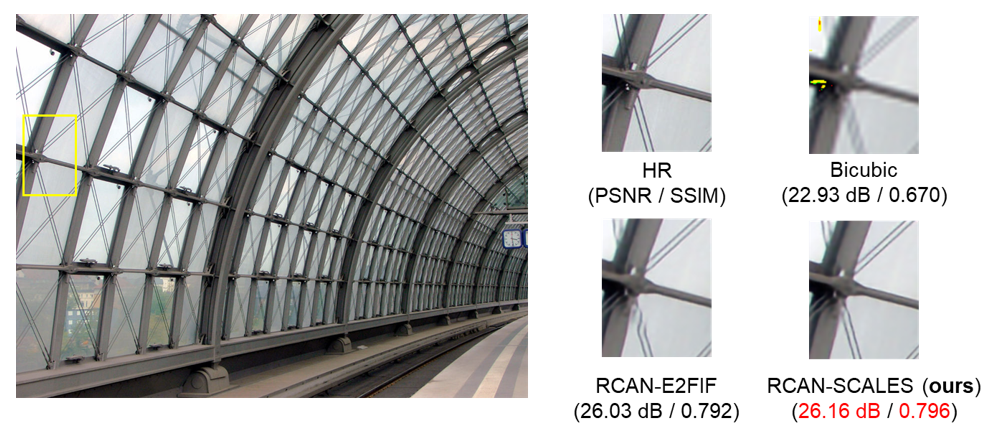}
  }
  
  \subfloat[Visual results from Set14 at $\times 2$ scale on EDSR architecture]{
    \label{subfig:visual_result_EDSR}
    \includegraphics[width=0.45\textwidth]{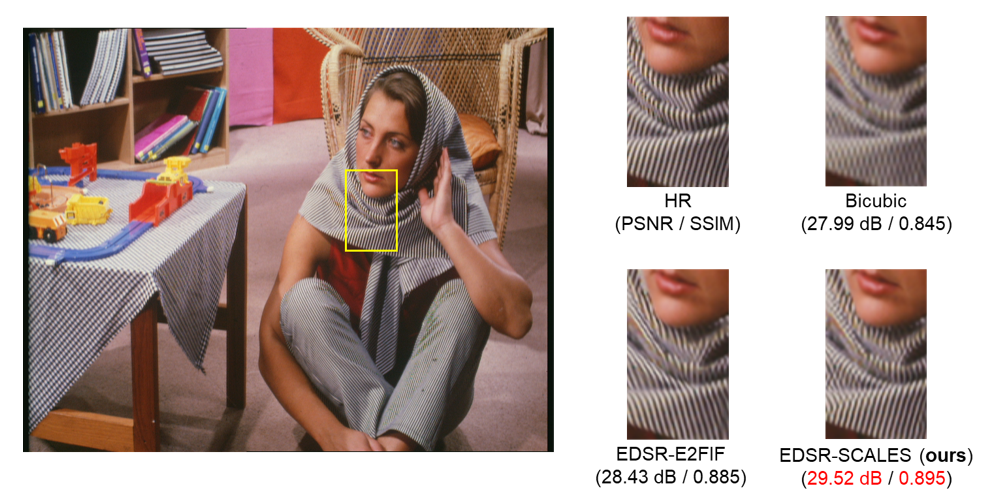}
  }
  \caption{
  Visual comparison of SCALES and the prior art method.
  }
  \label{fig: visual results}
  \vspace{-5pt}
\end{figure}

Qualitative results are shown in Figure~\ref{fig: visual results}.
We can observe that SCALES can alleviate the blurring artifacts
and reconstruct clearer images.
What's more, SCALES produces more faithful results to the ground truth,
compared to the prior art method E2FIF.
For example, in Fig.~\ref{subfig:visual_result_RCAN}, SCALES has no 
distortion and in Fig~\ref{subfig:visual_result_EDSR}, SCALES generates
the stripes with correct directions
while E2FIF fails to achieve this.

\subsection{Ablation Study}
\label{subsec: ablation}
We evaluate the effect of the proposed individual components on SRResNet in 
Table~\ref{tab:ablation}.
First, using layer-wise scaling factor (LSF) already outperforms
the prior art method E2FIF with fewer operations.
The computation cost is reduced because we remove BN in SRResNet-E2FIF.
The proposed channel-wise re-scaling method further improves PSNR by 0.12dB and 0.05dB on Set5 and Urban100, respectively, 
only increasing 4\% operations.
The spatial re-scaling method further achieves 0.18dB and 0.15dB improvement on the two datasets.
The increase in the number of parameters is negligible for all the components.


\begin{table}[!tb]
\centering
\caption{The effect of different components in SCALES.
OPs are calculated based on the 128×128 input image.
Note that the result of E2FIF is different from the original paper 
because we train all the model with RGB input instead of YCbCr input.
}
\label{tab:ablation}
\scalebox{0.85}{
\begin{tabular}{c|c|cc|cc}
\hline
\multirow{2}{*}{Method} & \multirow{2}{*}{OPs} & \multicolumn{2}{c|}{Set5} & \multicolumn{2}{c}{Urban100} \\ \cline{3-6}
                             &                                         & PSNR        & SSIM        & PSNR          & SSIM         \\ \hline
SRResNet-E2FIF           & 1.83G                                & 31.27       & 0.880        & 25.07         & 0.748        \\  \hdashline
LSF                    & 1.56G                                  & 31.30       & 0.880       & 25.09         & 0.751        \\
LSF + chl. re-scale & 1.63G                               & 31.42       & 0.880       & 25.14         & 0.753        \\
LSF + spatial re-scale  & 1.67G                                  & 31.48       & 0.882       & 25.24         & 0.756        \\
SCALES             & 1.74G                                & 31.54       & 0.883       & 25.27         & 0.757        \\ \hline
\end{tabular}
}
\end{table}





\subsection{Deployment Efficiency}
We compare the memory and computation cost in Table~\ref{tab:result_srresnet} and~\ref{tab:result_swinir_and_hat}.
The number of parameters and operations
are calculated following~\cite{zhou2016dorefa,liu2018bi}:
$OPs=OPs^f + OPs^b/64, Param=Param^f+Param^b/32$.
We evaluate $OPs$ on a $1280\times720$ HR image.
In Table~\ref{tab:result_srresnet},
SCALES has the smallest number of parameters and operations.
Compared with the prior art method E2FIF,
SCALES has better performance with 1K parameters and 0.3G operations reduction
due to the removal of BN.
In Table~\ref{tab:result_swinir_and_hat},
SCALES has approximately $10\times$ and $20\times$ parameter reduction compared to SwinIR-FP and HAT-FP, respectively.
Compared to the baseline, SCALES only introduces negligible parameters
while having significantly better performance.


\begin{table}[!tbp]
\centering
\caption{Inference latency on mobile phone.}
\label{tab:latency}
\scalebox{0.8}{
\begin{tabular}{c|ccc|cc|cc}
\hline
\multirow{2}{*}{SRResNet}            & \multirow{2}{*}{OPs} & \multirow{2}{*}{Params} & \multicolumn{1}{l|}{\multirow{2}{*}{Latency}} & \multicolumn{2}{c|}{Set14}                             & \multicolumn{2}{c}{B100}                              \\ \cline{5-8} 
                                     &                      &                         & \multicolumn{1}{l|}{}                         & PSNR                      & SSIM                       & PSNR                      & SSIM                      \\ \hline
FP SRResNet                                 & 64.98G               & 1.52M                   & 1649 ms                                        & 28.25                     & 0.773                      & 27.38                     & 0.727                     \\
E2FIF                                & 1.83G                & 0.03M                   & 197 ms                                          & 27.93                     & 0.766                      & 27.20                     & 0.723                     \\
\multicolumn{1}{l|}{SCALES (chl=64)} & 1.74G                & 0.03M                   & 237 ms                                           & \multicolumn{1}{l}{28.15} & \multicolumn{1}{l|}{0.770} & \multicolumn{1}{l}{27.28} & \multicolumn{1}{l}{0.726} \\
\multicolumn{1}{l|}{SCALES (chl=40)} & 0.83G                & 0.02M                   & 166 ms                                          & \multicolumn{1}{l}{28.02} & \multicolumn{1}{l|}{0.767} & \multicolumn{1}{l}{27.18} & \multicolumn{1}{l}{0.722} \\ \hline
\end{tabular}
}
\vspace{-10pt}
\end{table}


We also benchmark the latency of our method on 
Redmi K40S phone with a Qualcomm Snapdragon 870 SoC using Larq,
an open-source library for deploying BNNs.
We report the average latency of 100 times inference
using a single thread.
As shown in Table~\ref{tab:latency},
with our proposed SCALES (with the number of channels equal to 40),
we  can achieve $9.9\times$ speedup compared to the FP counterpart,
and $1.2\times$ speedup compared to the prior art method E2FIF with on-par performance.

\section{Conclusion}
\label{sec:conclusion}

In this paper, we propose an effective binarization method SCALES for both 
CNN-based and Transformer-based image SR networks,
based on our observation that activations in the SR network exhibit large pixel-to-pixel, channel-to-channel, layer-to-layer, and image-to-image variation.
With SCALES,
we improve the performance of binary CNN-based networks 
e.g., 0.2dB over the prior art method, with fewer parameters and operations.
We also achieve an accurate binary Transformer-based network for the first time,
attaining more than 1dB improvement over the baseline method.

\bibliographystyle{IEEEtran}
\bibliography{ref}

\end{document}